\documentclass[letterpaper, 10 pt, journal, twoside]{IEEEtran}
\usepackage[utf8]{inputenc}
\usepackage{url}
\usepackage{cite}
\usepackage{amsmath,amssymb,amsfonts}
\usepackage{textcomp}
\usepackage{lettrine}
\usepackage{threeparttable}
\usepackage{subcaption}
\usepackage[export]{adjustbox}
\newcommand{\subparagraph}{}
\usepackage{leftidx}
\usepackage{siunitx}
\usepackage{multirow}
\usepackage{booktabs}
\usepackage{dblfloatfix}
\usepackage{balance}
\usepackage{times}
\usepackage{siunitx}
\usepackage[table]{xcolor}
\newcommand{\ra}[1]{\renewcommand{\arraystretch}{#1}}
\usepackage[mathscr]{euscript}

\DeclareSymbolFont{rsfs}{U}{rsfs}{m}{n}
\DeclareSymbolFontAlphabet{\mathscrsfs}{rsfs}

\usepackage{hyperref}
\usepackage{algorithm2e}
\RestyleAlgo{ruled}
\usepackage{mathtools}
\usepackage{setspace}

\renewcommand{\arraystretch}{1.00}

\def\BibTeX{{\rm B\kern-.05em{\sc i\kern-.025em b}\kern-.08em
    T\kern-.1667em\lower.7ex\hbox{E}\kern-.125emX}}

\newtoggle{redtext}
\togglefalse{redtext} 

\begin{document}

\title{\LARGE \bf 
Tightly Coupled SLAM with Imprecise Architectural Plans
}

\author{Muhammad Shaheer$^1$, Jose Andres Millan-Romera$^1$, Hriday Bavle$^1$, Marco Giberna$^1$,\\ Jose Luis Sanchez-Lopez$^1$,  Javier Civera$^2$, and Holger Voos$^1$ 
\thanks{$^1$Authors are with the Automation and Robotics Research Group, Interdisciplinary Centre for Security, Reliability and Trust, University of Luxembourg. Holger Voos is also associated with the Faculty of Science, Technology and Medicine, University of Luxembourg, Luxembourg.\tt{\small{\{muhammad.shaheer, jose.millan, hriday.bavle, marco.giberna, joseluis.sanchezlopez, holger.voos\}}@uni.lu
}}
\thanks{$^2$Author is with I3A, Universidad de Zaragoza.
\tt{\small{jcivera@unizar.es}}}
\thanks{*This research was funded in whole, or in part, by the Luxembourg National Research Fund (FNR) under the projects 17097684/RoboSAUR and C22/IS/17387634/DEUS, by a partnership between the Interdisciplinary Center for Security Reliability and Trust (SnT) of the University of Luxembourg and Stugalux Construction S.A. and by the Spanish Government under Grants PID2021-127685NB-I00 and TED2021-131150B-I00.} 
\thanks{*For the purpose of open access, and in fulfilment of the obligations arising from the grant agreement, the author has applied a Creative Commons Attribution 4.0 International (CC BY 4.0) license to any  Author Accepted Manuscript version arising from this submission.}
}
\maketitle

\begin{abstract}
\label{abstract}
Robots navigating indoor environments often have access to architectural plans, which can serve as prior knowledge to enhance their localization and mapping capabilities.
While some SLAM algorithms leverage these plans for global localization in real-world environments, they typically overlook a critical challenge: the ``as-planned" architectural designs frequently deviate from the ``as-built" real-world environments.
To address this gap, we present a novel algorithm that tightly couples LIDAR-based simultaneous localization and mapping with architectural plans in the presence of deviations.
Our method utilizes a multi-layered semantic representation to not only  
localize the robot, but also to estimate global alignment and structural deviations between ``as-planned" and ``as-built" environments in real-time.
%
To validate our approach, we performed experiments in simulated and real datasets demonstrating robustness to structural deviations up to  $35$ cm and $15^\circ$.
On average, our method achieves $43\%$ less localization error than baselines in simulated environments, while in real environments, the ``as-built" 3D maps show $7\%$ lower average alignment error.

\noindent \textbf{Paper Video:} \href{https://www.youtube.com/watch?v=9O0qwNhTuqk}{https://www.youtube.com/watch?v=9O0qwNhTuqk}
\end{abstract}

\section{Introduction}
\label{introduction}
\IEEEPARstart{P}{rior} information from architectural plans can be used to enhance the localization and mapping accuracy of mobile robots. 
However, real-world buildings rarely match their plans perfectly due to construction tolerances and modifications \cite{jenewein2019validation, JIA2022104096}. Incorporating such imprecise prior information into robot navigation pipelines can introduce systematic errors, potentially damaging the localization and mapping accuracy, instead of helping to improve it.
%
To address this issue, it is necessary to tightly couple the Simultaneous Localization and Mapping (SLAM) pipeline with architectural plans. This requires a unified representation of both the ``as-planned'' and ``as-built'' environments --one that captures not only the geometry but also architectural semantics. Such a representation will first enable the matching of structural elements between the two instances, and subsequently, the estimation of their deviations.
\begin{figure}[!t] 
  \centering
  \includegraphics[width=0.95\columnwidth]{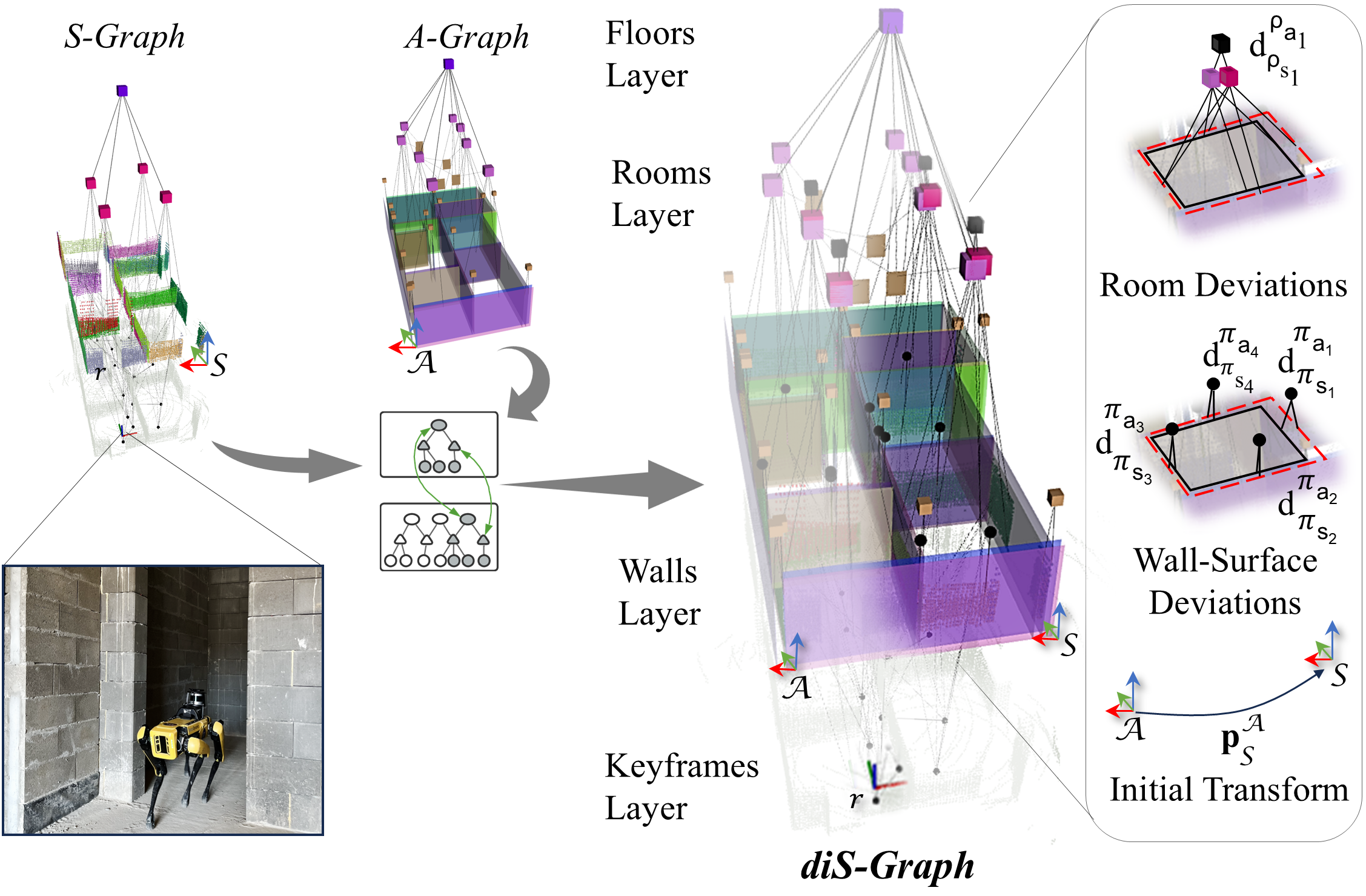}
  \caption{\textbf{\textit{diS-Graphs} overview.} Our method 
  couples a hierarchical SLAM factor graph built online by a robot, with an architectural plan (also modeled as a hierarchical factor graph) that may contain deviations, 
  to form
  a deviations-informed Situational Graph (\textit{diS-Graph}, center).
  The zoomed-in view (right) illustrates 
  that this coupling enables the estimation of
  the rigid transformation $ \mathbf{p}^\mathscr{A}_{\mathcal{S}}$ between 
  both graphs
  and, additionally, the wall-surface and room deviations $\mathbf{d}^{\pi_{a_i}}_{\pi_{s_i}}$ and $\mathbf{d}^{\rho_{a_j}}_{\rho_{s_j}}$ respectively.}
  \label{fig:wideimage}
\end{figure}
Although there are various environment representation techniques such as \textit{occupancy grids} \cite{occupancy_g_map},
none of them explicitly models the semantic and hierarchical information of the environment, which is needed to identify the deviated structural elements.
Recent approaches such as 3D scene graphs  \cite{armeni_3d_2019,hughes_hydra_2022} or Situational Graphs (\textit{S-Graphs}) \cite{bavle_situational_2022, bavle_sgraphs_2023}, represent a robot’s environment in a compact and hierarchical manner, encoding high-level semantic abstractions (for example, walls and rooms) and their relationships (e.g., a set of walls forms a room). Herein, \textit{S-Graphs} extend 3D scene graphs by merging geometric models of the environment generated by SLAM approaches with 3D scene graphs into a multi-layered jointly optimizable factor graph.
This representation, combined with the prior information extracted from architectural plans, can be used to provide fast and efficient localization.

Informed \textit{S-Graphs} (\textit{iS-Graphs})~\cite{shaheer_graph-based_2023} further extend \textit{S-Graphs} by using architectural plans to provide accurate localization over the resulting hierarchical factor graphs.
 However, its success is based on the assumption that there are no deviations between the ``as-built'' and the ``as-planned''. In reality, this is never the case, and the building elements exhibit certain deviations with respect to their planned geometries.

The main contribution of this paper is a novel method capable of coupling architectural plans (``as-planned'') and SLAM (``as-built'') data even in the presence of deviations, as shown in Fig. \ref{fig:wideimage}. We call this algorithm \textit{Deviations Informed Situational Graphs} or \textit{diS-Graphs} in short. 

In summary, \textit{diS-Graphs} performs three interconnected tasks simultaneously:
\begin{itemize}
     \item 
    Tightly coupling the SLAM factor graph with architectural plans, in order to match structural elements (\textit{walls}, \textit{rooms}) between planned and built environments.
    \item 
    Globally localizing the robot in imprecise architectural plans that may contain deviations from the real building.
    \item 
    Detecting and estimating the deviations between the matched structural elements of ``as-planned'' and ``as-built'' environments in real-time.
\end{itemize}


\section{Related Works}
\label{related_works}
Most localization techniques using prior information from architectural plans assume that the environments are built precisely according to the plans. One of the most commonly used localization techniques in 2D metric prior maps is Monte Carlo Localization (MCL) \cite{dieter_fox_monte_2001}, \cite{shaheer_robot_2023} but it is not scalable to large-scale complex environments. Boniardi et al \cite{boniardi_pose_2019} use a technique that scales to more complex environments by aligning a scan-based map with CAD-based floor plans. OGM2PGM \cite{torres_occupancy_2023} also scales to larger environments by converting the 2D floor plan to an occupancy grid map (OGM) and using a pose-graph map (PGM)  to localize the robot. UKFL \cite{koide_portable_2019} further enhances the localization accuracy using an unscented Kalman filter to localize the robot in 3D metric meshes. Recent techniques such as \cite{kuang_ir-mcl_2023} exploit neural networks to localize the robot using an implicit neural representation of the floor plans. 
These techniques rely solely on geometric information from architectural plans, ignoring available semantic data (room types, door locations, functional spaces) that could disambiguate similar areas and improve localization. Additionally, these geometric-only approaches are vulnerable to floor plan inaccuracies or outdated information.

To address these limitations,
semantic-based localization techniques, such as Mendez et al.~\cite{mendez_sedar_2018} use semantic cues from architectural plans and sensor information to improve localization accuracy. Boniardi et al.~\cite{boniardi_robot_2019} exploit the semantics of the room in architectural plans to do robot localization by matching the detected rooms from sensor data. Wang et al.~\cite{wang_glfp_2019} leverage prelabeled architectural features, such as wall intersections and corners, as landmarks in floor plans, and match them with detection from sensor data to jointly perform mapping and localization. Zimmerman et al.~\cite{zimmerman_long-term_2023}, ~\cite{zimmerman_constructing_2023} use high-level semantic information in floor plans, derived from object detection, along with geometric data from 2D LiDAR to perform long-term robot localization in floor plans. Huan et al. \cite{BIM-LOC} convert architectural plans into semantically enriched point cloud maps, followed by a coarse-to-fine localization process using ICP. Gao et al. \cite{gao_fp-loc_2022} used neural networks to detect vertical elements from floor plans to do LiDAR based localization.
These methods are prone to inaccuracies due to misidentification and errors in the pose estimate of semantic elements.  Moreover, they treat semantic elements in isolation, failing to leverage the rich contextual information embedded in their spatial and functional relationships. Recent work, such as Shaheer et al. \cite{shaheer_graph-based_2023} exploits the topological relationship between semantic elements to localize the robot with respect to architectural plans.
However, all the above mentioned approaches assume no deviations between the architectural plans and the actual environment. 

Some recent works leverage imprecise floor plans for localization.
Boniardi et al.~\cite{boniardi_robust_2017} integrate localization techniques to take advantage of the information embedded in the CAD drawing, and the real-world observations acquired during navigation, which may not be reflected in the floor plan. Li et al.~\cite{li_online_2020} presented a 2D LiDAR-based localization system in imprecise floor plans using stochastic gradient descent (SGD) with a scan matching algorithm. Chan et al. \cite{li_partial-map-based_2021} presented a 2D LiDAR-based localization in floor plans that integrates SLAM with MCL. Blum et al. \cite{blum_precise_2021} use neural networks for feature segmentation and combine them with LiDAR data for localization in imprecise floor plans.
%
Despite advances in robot localization using imprecise floor plans, no existing approach combines localization with the estimation of structural deviations between ``as-planned'' and ``as-built'' environments, limiting the ability to assess construction accuracy in real time. In this work, we address this limitation by presenting a unified framework that enables both global localization and deviation estimation.
\begin{figure*}[ht]
    \centering
    \includegraphics[width=0.9\textwidth]{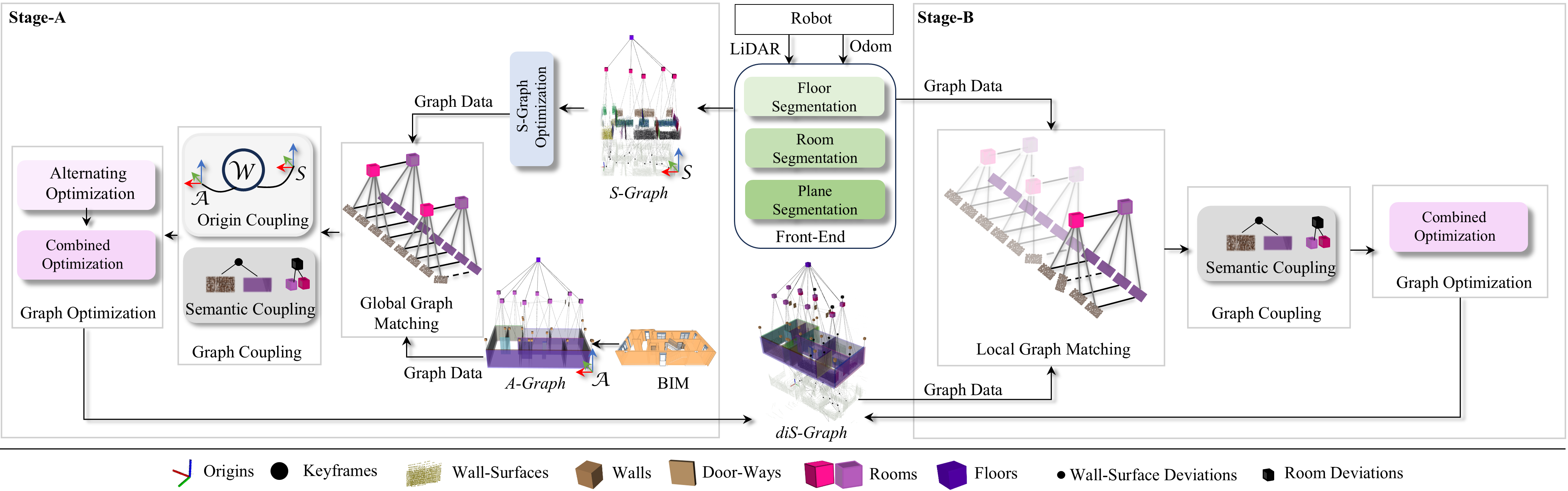}
    \caption{\textbf{System Architecture.} The inputs to our method are an \textit{A-Graph} generated from an architectural plan, and an \textit{S-Graph} estimated online from the 3D LiDAR and the robot odometry. Stage-A is run first, and only once, for initial estimates of global localization and deviation. Once Stage-A is successful, Stage-B is run sequentially to match, couple, and optimize newly incorporated observations incrementally.
    }
    \label{fig:system_architecture}
\end{figure*}
\section{
System Architecture
}
\label{sec:proposed_approach}
\begin{figure}[!t]
    \centering
    \includegraphics[width=0.40\textwidth]{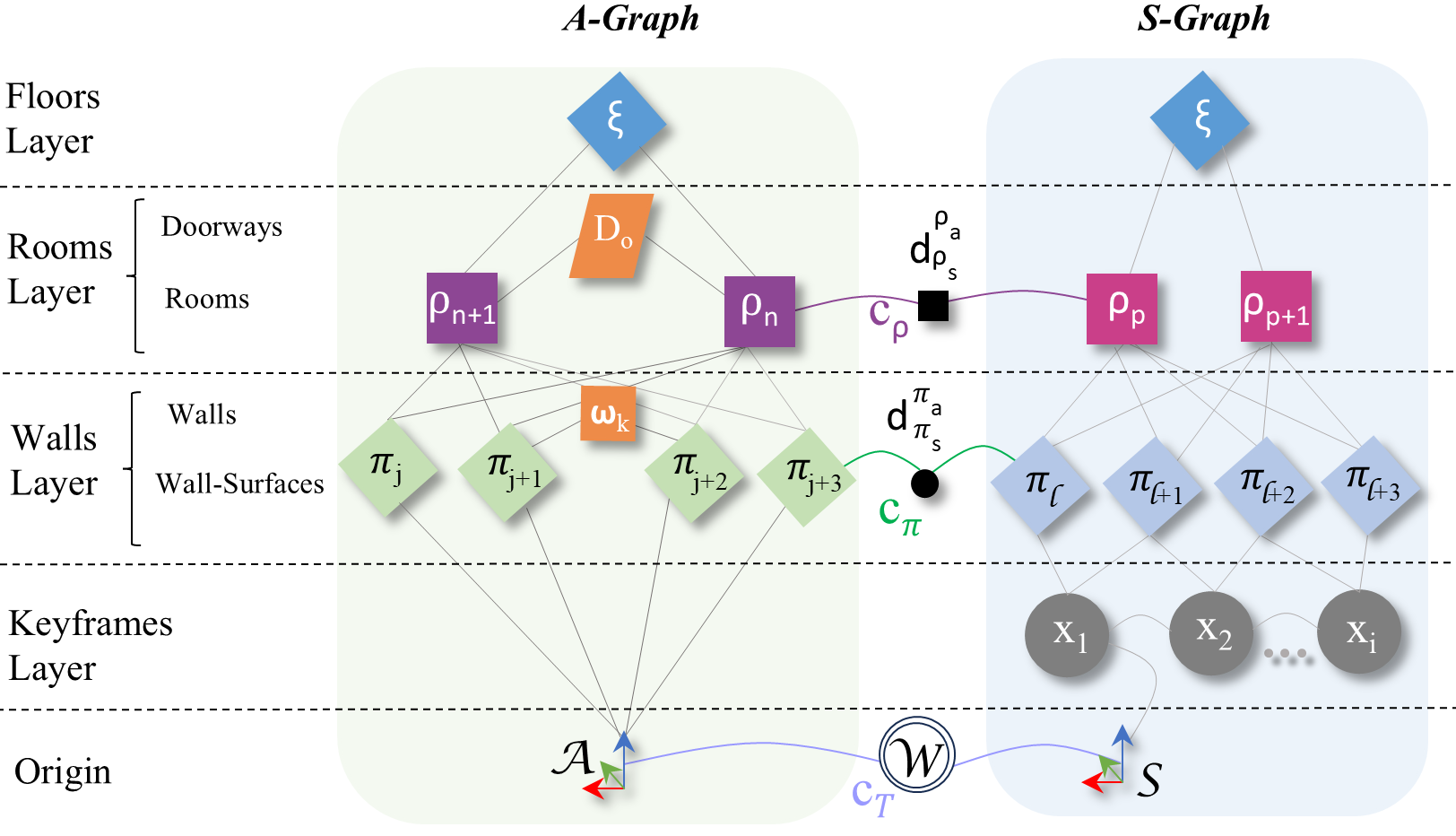}
        \caption{\textbf{\textit{diS-Graph} structure,} 
        tightly coupling an \textit{A-Graph} and a \textit{S-Graph}. ${\boldsymbol{\mathbf{d}}^{{\boldsymbol{\rho_a}}}_{{\boldsymbol{\rho_s}}}}$ and ${\boldsymbol{\mathbf{d}}^{{\boldsymbol{\pi_a}}}_{{\boldsymbol{\pi_s}}}}$ refer to the deviation nodes between rooms and wall-surfaces respectively, with $c_{\mathbf{d}_{\boldsymbol{\rho}}}$ and $c_{\mathbf{d}_{\boldsymbol{\pi}}}$ as their associated cost functions. $\mathscr{W}$ is the anchor node, and, $c_T$ is the associated cost function to estimate the transformation $\mathbf{p}^{\mathscr{A}}_{\mathcal{S}}$ between the two graph origins $\mathscr{A}$ and $\mathcal{S}$.
        }
    \label{fig:graph_structure}
\end{figure}
We propose an algorithm that tightly couples SLAM and architectural plans and jointly optimizes them.
\subsection{Factor Graph Structure and Notation}

Our algorithm represents the SLAM backend and the architectural plans using hierarchical factor graphs (see Fig. \ref{fig:graph_structure}). For clarity, we first introduce here the terminology and notation used throughout this paper.

\textbf{Architectural Graph (\textit{A-Graph}).} 
Three-layered hierarchical factor graph containing the geometry, semantics, and topology of an environment, generated from its architectural plan \cite{shaheer_graph-based_2023}. It models the environment ``as-planned'' by the architect.

\textbf{Situational Graph (\textit{S-Graph}).} 
 Four-layered hierarchical optimizable factor graph, estimated online from 3D LiDAR and odometry measurements, that models the ``as-built'' environment \cite{bavle_situational_2022}, \cite{bavle_sgraphs_2023}. It also includes keyframes in addition to the geometry, semantics, and topology of the environment.

\textbf{Deviations Informed Situational Graph (\textit{diS-Graph}).}
Joint graph, coupling both the \textit{A-Graph} and the \textit{S-Graph} of a building.
The layers of these graphs are as follows.

\textbf{\textit{Origins Layer.}}

The \textit{S-Graph} and \textit{A-Graph} are defined in their respective frames ${\mathcal{S}}$ and ${\mathscr{A}}$, with anchor frame ${\mathscr{W}}$ set such that $\mathbf{p}^{\mathscr{A}}_\mathscr{W} = \mathbf{0}$.

\textbf{\textit{Keyframes Layer.}}

Exclusive to \textit{S-Graphs}, this layer contains robot poses in the ${\mathcal{S}}$ frame: $\mathbf{x}^\mathcal{S}_{{r}_i} \in SE(3) \ \forall \ i$.

\textbf{\textit{Walls Layer.}} 
In \textit{A-Graph}: Contains wall-surfaces ${\boldsymbol{\pi}}^\mathscr{A}_{a_j} \in S(2)\times \mathbb{R} \ \forall \ j$ and walls $\boldsymbol{\omega}^\mathscr{A}_{a_k} \in SE(3) \ \forall \ k$. Wall-surfaces are defined by their normals and distances to the origin \cite{bavle_situational_2022}. A wall is constrained by two opposite wall-surfaces via the factor $ \|f({{\boldsymbol{\pi}}}^\mathscr{A}_{a_1}, {{\boldsymbol{\pi}}}^\mathscr{A}_{a_2}, \mathbf{s^\mathscr{A}_\omega})\|^2_\Lambda$, defined in Eq. 2 of \cite{shaheer_graph-based_2023}. 
In \textit{S-Graph}: Contains only wall-surfaces ${\boldsymbol{\pi}}^\mathcal{S}_{s_l} \in S(2)\times \mathbb{R} \ \forall \ l$ extracted from 3D LiDAR scans and linked to keyframes via pose-plane constraints.
We couple both graphs in our \textit{diS-Graphs} via wall deviation nodes $\mathbf{d}^{{\boldsymbol{\pi}}_{a_i}}_{{\boldsymbol{\pi}}_{s_i}} $ and their associated factors $c_{{\boldsymbol{\pi}}_m} \ \forall \ m$, that we show in Fig. \ref{fig:graph_structure} and define in Eq. \ref{eq:wall_association_cf} in this paper.

\textbf{\textit{Rooms Layer.}} 
In \textit{A-Graph}: Contains rooms ${\boldsymbol{\rho}}^\mathscr{A}_{a_n} \in SE(3) \ \forall \ n$ composed of four wall-surfaces, and doorways $\boldsymbol{\mathcal{D}}^\mathcal{A}_{a_o} \in SE(3) \ \forall \ o$. A doorway constrains two rooms via the factor $\|f({{\boldsymbol{\rho}}}^\mathscr{A}_{a_1}, {\boldsymbol{\mathcal{D}}}^{{\boldsymbol{\rho}}_1}_{a_1}) - f({{\boldsymbol{\rho}}}^\mathscr{A}_{a_2}, {\boldsymbol{\mathcal{D}}}^{{\boldsymbol{\rho}}_2}_{a_2})\|^2_\Lambda$, that is defined in Eq. 3 of \cite{shaheer_graph-based_2023}. A room is constrained by four wall-surfaces via factor $ f({{\boldsymbol{\pi}}}^\mathscr{A}_{x_{1}}, {{\boldsymbol{\pi}}}^\mathscr{A}_{x_{2}}, {{\boldsymbol{\pi}}}^\mathscr{A}_{y_{1}}, {{\boldsymbol{\pi}}}^\mathscr{A}_{y_{2}})\|^2_\Lambda$, defined in Eq. 7 of \cite{bavle_sgraphs_2023}.
In \textit{S-Graph}: Contains rooms of four wall-surfaces ${\boldsymbol{\rho}}^\mathcal{S}_{s_p} \in SE(3) \ \forall \ p$ or two wall-surfaces ${\boldsymbol{\kappa}}_{s_q} \in SE(3) \ \forall \ q$.
Both graphs are connected by room deviation factors $\mathbf{d}^{{\boldsymbol{\rho}}_{a_i}}_{{\boldsymbol{\rho}}_{s_i}}$ in our \textit{diS-Graph}.

\textbf{\textit{Floors Layer.}} In both \textit{A-Graph} and \textit{S-Graph} this layer consists of a floor center node represented as $\boldsymbol{\xi}^\mathscr{A}_{s}  \in \mathit{SE}(3)$ and $\boldsymbol{\xi}^\mathcal{S}_{s}  \in \mathit{SE}(3) $ respectively, constraining all rooms present at that particular floor level with the factor $ \|f({\xi}^{\mathscr{A}, \mathcal{S}}_{a,s}, {{\boldsymbol{\rho}}}^{\mathscr{A}, \mathcal{S}}_{a_{n},s_{p}})\|^2_\Lambda$, defined in Eq. 9 of the reference \cite{bavle_sgraphs_2023}.

More details on the type of factors between the different elements of the graphs can be found in \cite{bavle_sgraphs_2023} \cite{shaheer_graph-based_2023}.
\subsection{The Algorithm}
Our algorithm has two stages, as shown in Fig.~\ref{fig:system_architecture}. 
In Stage-A, the \textit{A-Graph} is first matched and 
coupled
with the \textit{S-Graph}, and then 
jointly
optimized. Stage-A runs until it finds the initial match between the two graphs, and provides the initial estimates of both the transformation between graphs and potential deviations between their elements. 
Afterwards, in Stage-B, the joint graph is continuously refined through incremental matching and optimization as the robot navigates the environment, incorporating newly detected semantic entities. This tight coupling enables continuous refinement of both the robot's localization and the detected structural deviations during navigation.
Both stages of our algorithm consist of multiple processes, introduced briefly here and fully detailed later in their respective sections. 

\textbf{Graph Matching (Section \ref{sec:graph_matching}).}
The Global Graph Matching in Stage-A provides 
the first unique match, when there is sufficient information, between the \textit{S-Graph} and \textit{A-Graph} at room and wall-surface levels, accounting for potential deviations.
In Stage-B, the Local Graph Matching extends the previously matched elements in the \textit{diS-Graph} with newly detected elements following an incremental approach.

\textbf{Graph Coupling (Section.~\ref{sec:data_association}).}
It is performed for the candidates matched by graph matching. In Stage-A, graph coupling registers the origins of two graphs along with the semantic 
coupling (wall-surfaces and rooms), and explicit mapping of the deviation factors. Stage-B only performs the semantic coupling between newly observed entities and their correspondences in the plan, with explicit deviation factors. 

\textbf{Graph Optimization (Section.~\ref{sec:graph_optimization}).}
It optimizes the coupled \textit{diS-Graph}.
Stage-A involves two steps: alternating optimization and joint optimization. Alternating optimization estimates two types of variables, 1) The transformation between the origins of the \textit{S-Graph} and the \textit{A-Graph}, yielding global localization, and 2) the potential deviations between the matched graph entities. Stage-B jointly optimizes the \textit{diS-Graph} from initial seeds for the global transformation and map deviations. 

\section{Graph Matching}
\label{sec:graph_matching}
Our graph matching extends the method presented in \cite{shaheer_graph-based_2023}.
In \cite{shaheer_graph-based_2023}, a top-down potential candidate search between an \textit{A-Graph} and a \textit{S-Graph} is performed by leveraging their hierarchical structures.
To assess the overall consistency of matching candidate, two verification steps are applied iteratively. First, the consistency of the node type and graph structure is verified. And secondly, geometric consistency is checked over a certain predefined threshold \cite{lusk2021clipper}. Finally, the consistencies of all candidates are compared, and the candidate with the best consistency is selected as the final unique match.
Note that symmetries may affect our algorithm. Their occurrence may result in false positives when the robot has only partially explored the environment and observed one of the symmetric places. They may also cause several equally good candidates if the robot has explored a larger fraction of the building.
\begin{figure}[h]
    \centering
    \includegraphics[width=0.40\textwidth]{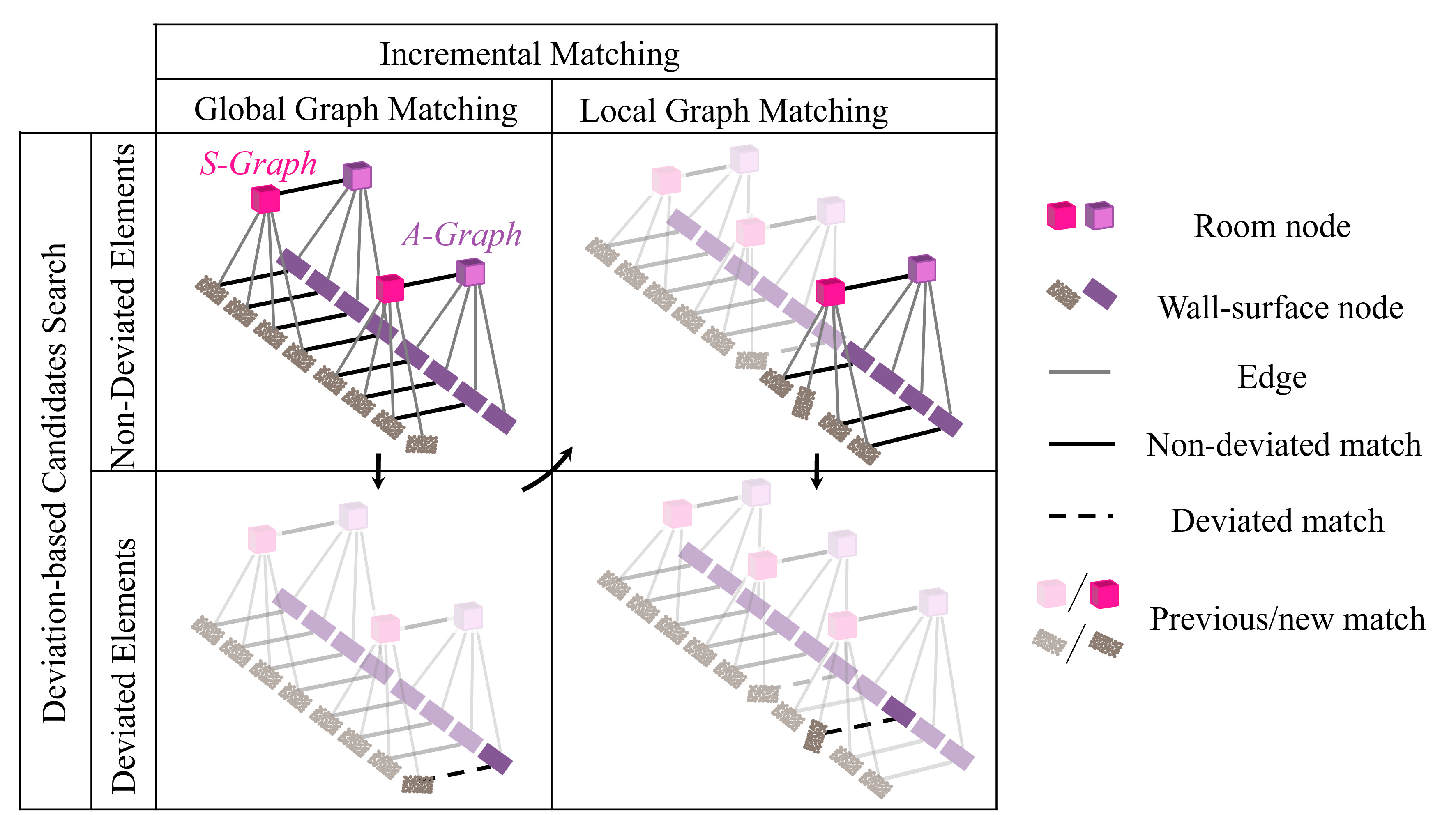}
        \caption{
        \textbf{Illustration of the graph matching workflow} between the semantic nodes of an \textit{A-Graph} and a \textit{S-Graph}. 
        }
    \label{fig:graph_matching_workflow}
\end{figure}
\subsection{Deviation-based Candidates Search}
\label{subsec:deviations}
The presence of deviations generates geometric inconsistencies that affect the aforementioned candidates' checks of \cite{shaheer_graph-based_2023}.
Concretely, a deviation in the position of a wall-surface implies a slight deviation in the center of its parent room as well. 
To handle this, we propose a two-stage search method in this paper, searching first for non-deviated wall-surfaces and including after that those elements that are deviated. 

\textbf{Non-deviated wall-surfaces search} To handle potential deviations, we tighten the matching criteria.
Specifically, during the generation of non-deviated \textit{room}-to-\textit{room} and \textit{room}-to-\textit{wall-surface} match candidates, we look for the candidates with higher consistency at each level.

\textbf{Deviated wall-surfaces detection} 
To identify deviated wall-surfaces that were not matched in the first stage but are connected to already. matched rooms, we now realx the matching criteria at \textit{wall-surface}-to-\textit{wall-surface} level to account for inconsistencies due to deviations.
To further speed up the candidate search, we incorporate the following additional information.

\textbf{\textit{{Orphan wall-surfaces:}}} We utilize wall-surfaces in the \textit{S-Graph} without a parent room for the assessment of the geometrical consistency of the final match candidates at the wall-surface level. 

\textbf{\textit{{Ground orientation:}}}
We exploit the ground plane normal in the~\textit{A-Graph} and the~\textit{S-Graph}, only allowing candidates with z-axis rotations.
Finally, the geometric consistency score provides a quantification of the probability of the deviation for each room and wall-surface, which is further used in the graph coupling
stage (Section \ref{sec:data_association}).
\subsection{Incremental Matching}
\label{subsec:incremental_matching}
To enhance the efficiency of the Graph Matching algorithm in~\cite{shaheer_graph-based_2023}, we implement an incremental approach with two stages (associated with stages A and B of the systems architecture of Fig.~\ref{fig:system_architecture}), namely \textbf{Global Graph Matching} (Fig.~\ref{fig:graph_matching_workflow} left) and \textbf{Local Graph Matching} (Fig.~\ref{fig:graph_matching_workflow} right), each of them based on the two deviation-based candidate search stages described in Section \ref{subsec:deviations}.
Until a first unique match has been found, the {Global Graph Matching} is executed for every new observation in the \textit{S-Graph}.
Afterwards, the {Local Graph Matching} is run every time the \textit{diS-Graph} is updated with newly observed rooms and wall-surfaces.
The graph elements that have been already matched are excluded from candidate generation here, and each assessment of intra-level consistency considers the previously matched elements at the corresponding level.
\section{Graph Coupling}
\label{sec:data_association}
\textbf{Origin Coupling.} 
We couple the two graphs using the anchor node $\mathscr{W}$ as a common reference frame, as shown in Fig.~\ref{fig:graph_structure}. This allows us to align both graphs in the frame $\mathscr{W}$ by estimating the transformation $\mathbf{p}^{\mathscr{A}}_{\mathcal{S}}$ between  $\mathscr{A}$ and $\mathcal{S}$ . The associated constraint $c_T$  is defined as:
\begin{equation}
c_T = \|{}\mathbf{p}^{\mathscr{A}}_{\mathcal{S}} \ominus \mathbf{p}^{\mathscr{W}}_{\mathscr{S}} \oplus \mathbf{p}^{\mathcal{W}}_{\mathscr{A}}  {}\|^2_{\Lambda_T}
\label{eq:origin_origin_tf_eq}
\end{equation}
${\mathbf{\Lambda}_{\boldsymbol{\tilde{\text{\textrm{T}}}}}}$ stands for the covariance of the cost, and it is assigned a high value to estimate the transformation accurately. Note that, since our approach simultaneously estimates both the global transformation and structural deviations, special care should be put in choosing the initial seed for this transformation and avoiding local minima coming from the combined effect of wall deviations and local registration.

\textbf{Semantic Entities Coupling.}
To estimate deviations between the wall-surfaces and rooms of the \textit{A-Graph} and the \textit{S-Graph}, we couple them with deviation nodes $\mathbf{d}^{{\boldsymbol{\pi}}_{a_i}}_{{\boldsymbol{\pi}}_{s_i}}$ and $\mathbf{d}^{{\boldsymbol{\rho}}_{a_i}}_{{\boldsymbol{\rho}}_{s_i}}$ respectively, as shown in Fig. \ref{fig:graph_structure}. 

\textit{\textbf{Room Coupling:}}
To estimate the deviation $\mathbf{d}^{{\boldsymbol{\rho}}_{a_i}}_{{\boldsymbol{\rho}}_{s_i}}$ in the pose of a pair of matched rooms of the two graphs, we define the cost function $c_{\mathbf{d}_{{\boldsymbol{\rho}}_i}}$ as:
\begin{equation}
c_{\mathbf{d}_{{\boldsymbol{\rho}}_i}} = \|\mathbf{d}^{{\boldsymbol{\rho}}_{a_i}}_{{\boldsymbol{\rho}}_{s_i}} \ominus {\boldsymbol{\rho}}^{\mathscr{W}}_{s_i} \oplus {\boldsymbol{\rho}}^{\mathscr{W}}_{a_i} \|^2_{\Lambda_{\mathbf{d}_{{\boldsymbol{\rho}}_i}}}
\label{eq:room_deva-cf}
\end{equation}
Here ${\boldsymbol{\rho}}^{\mathscr{W}}_{a_i}$ and ${\boldsymbol{\rho}}^{\mathscr{W}}_{s_i}$ are the room poses of \textit{A-Graph} and \textit{S-Graph} respectively, in common frame $\mathscr{W}$. $\mathbf{\Lambda}_{\boldsymbol{\tilde{\text{\textrm{d}}}}_{{\boldsymbol{{\boldsymbol{\rho}}_i}}}}$ is the covariance associated with the cost function depending on the probability of deviation estimated by the matching of the graph (Section \ref{sec:graph_matching}). Rooms with a higher probability of deviation assigned by graph matching have a higher covariance assigned to their cost function than rooms with lower deviation probability. If all matched rooms have the same deviation probability, they are assigned lower uniform covariances.

 \textit{\textbf{Wall-Surface 
 Coupling:
 }}
To estimate the deviation $\mathbf{d}^{{\boldsymbol{\pi}}_{a_i}}_{{\boldsymbol{\pi}}_{s_i}}$ between a pair of matched wall-surfaces of both graphs,  we define the cost function of deviation factor between two wall-surfaces $c_{\mathbf{d}_{\boldsymbol{\pi}}}$ as:
%
\begin{equation}
c_{\mathbf{d}_{{\boldsymbol{\pi}}_i}} = \|\mathbf{d}^{{\boldsymbol{\pi}}_{a_i}}_{{\boldsymbol{\pi}}_{s_i}} \ominus {\boldsymbol{\pi}}^{\mathscr{W}}_{s_i} \oplus {\boldsymbol{\pi}}^{\mathscr{W}}_{a_i} \|^2_{\Lambda_{\mathbf{d}_{{\boldsymbol{\pi}}_i}}}
\label{eq:wall_association_cf}
\end{equation}
Here ${\boldsymbol{\pi}}^{\mathscr{W}}_{a_i}$ and ${\boldsymbol{\pi}}^{\mathscr{W}}_{s_i}$ are the poses of wall-surfaces of \textit{A-Graph} and \textit{S-Graph} respectively, in common frame $\mathscr{W}$.\footnote{For compactness, we slightly abused notation by overloading the $\oplus$ and $\ominus$ operators. In Eq. \ref{eq:room_deva-cf}, they denote the pose composition and pose difference, respectively, and, in Eq. \ref{eq:wall_association_cf}, they denote the plane compositions and difference, respectively.} $\mathbf{\Lambda}_{\boldsymbol{\tilde{\text{\textrm{d}}}}_{{\boldsymbol{\pi}}_i}}$ is the covariance associated with the cost function.
Like rooms, wall-surfaces with a higher deviation are assigned higher covariances, and the ones with a lower deviation are assigned lower uniform covariances.
%




\begin{figure*}[h]
\centering
\begin{subfigure}{0.25\textwidth}
\raggedright
\includegraphics[width=1\textwidth]{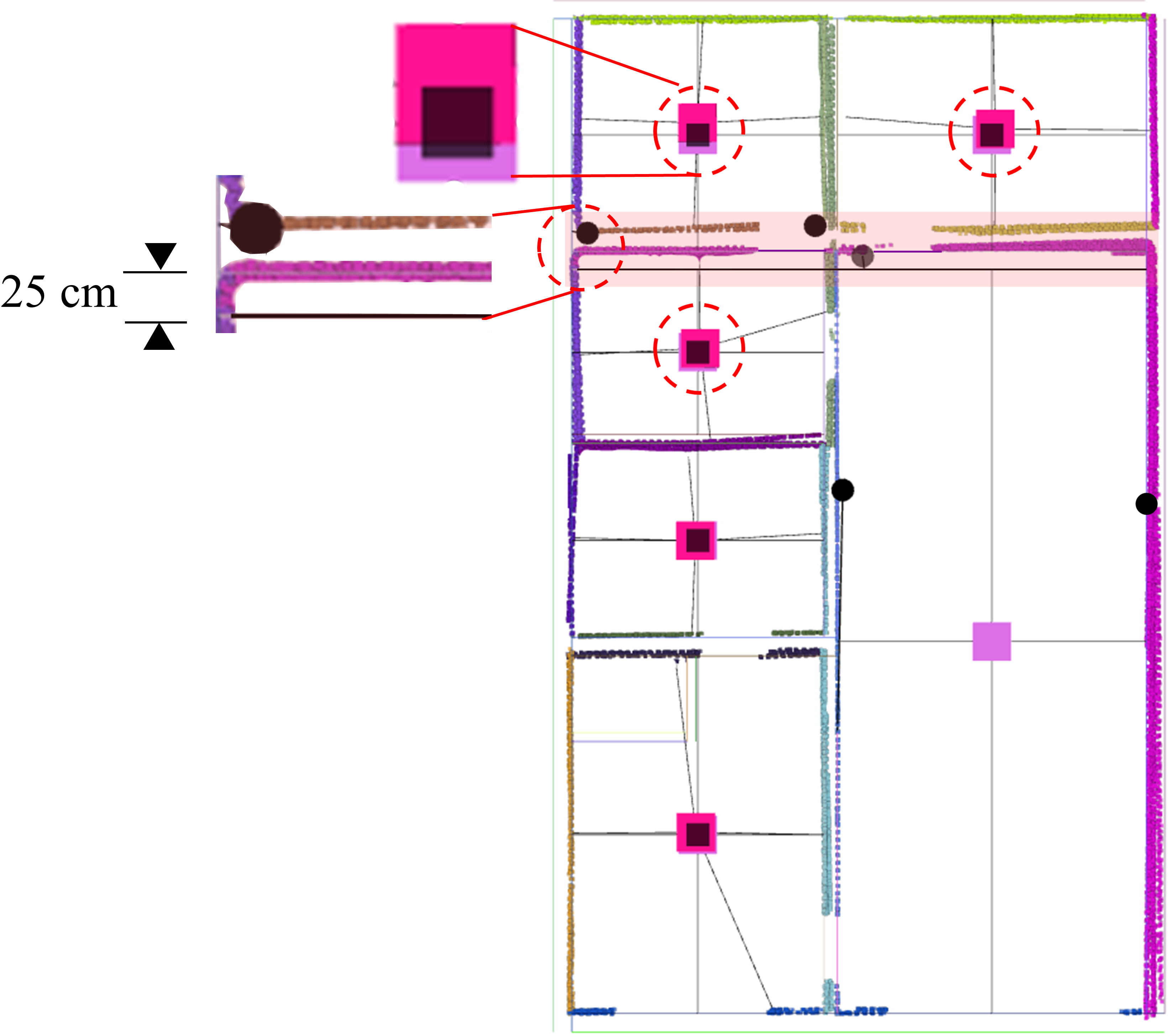}
\caption{RE1}
\label{fig:RE1}
\end{subfigure}
\begin{subfigure}{0.25\textwidth}
\raggedleft
\includegraphics[width=0.65\textwidth]{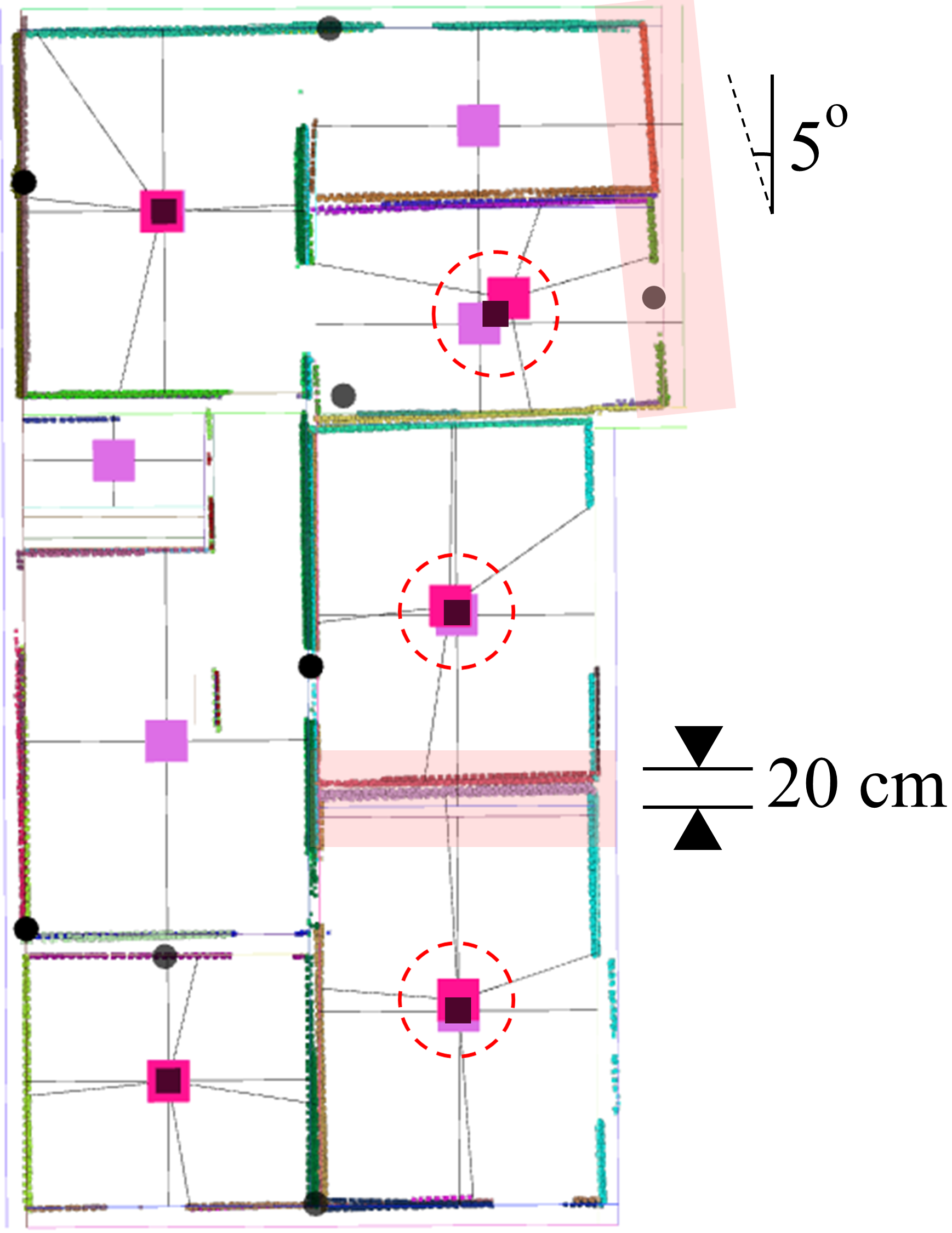}
\caption{RE2}
\label{fig:RE2}
\end{subfigure}
\begin{subfigure}{0.2\textwidth}
\raggedleft
\includegraphics[width=0.75\textwidth]{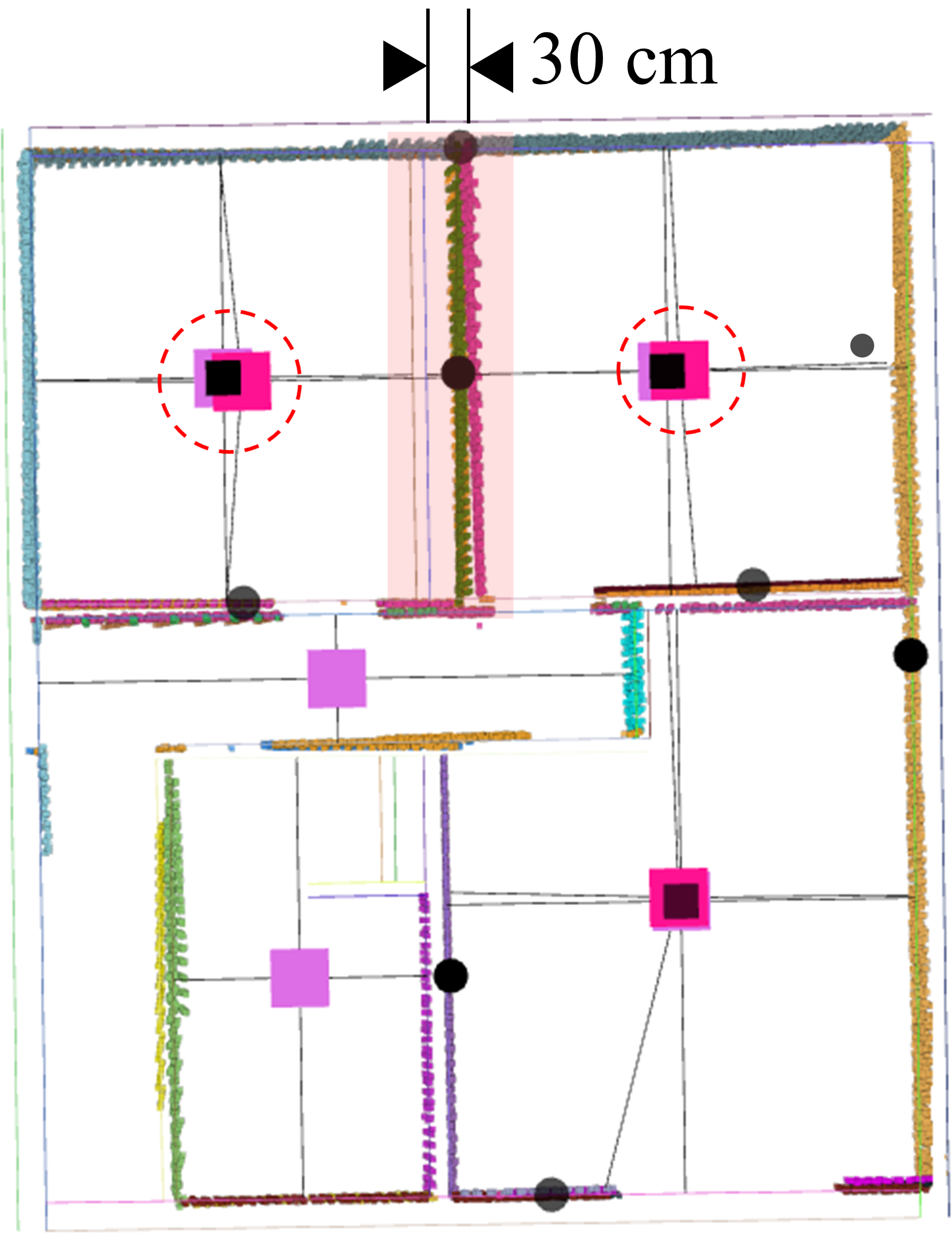}
\caption{RE3}
\end{subfigure}
\begin{subfigure}{0.2\textwidth}
\raggedleft
\includegraphics[width=0.9\textwidth]{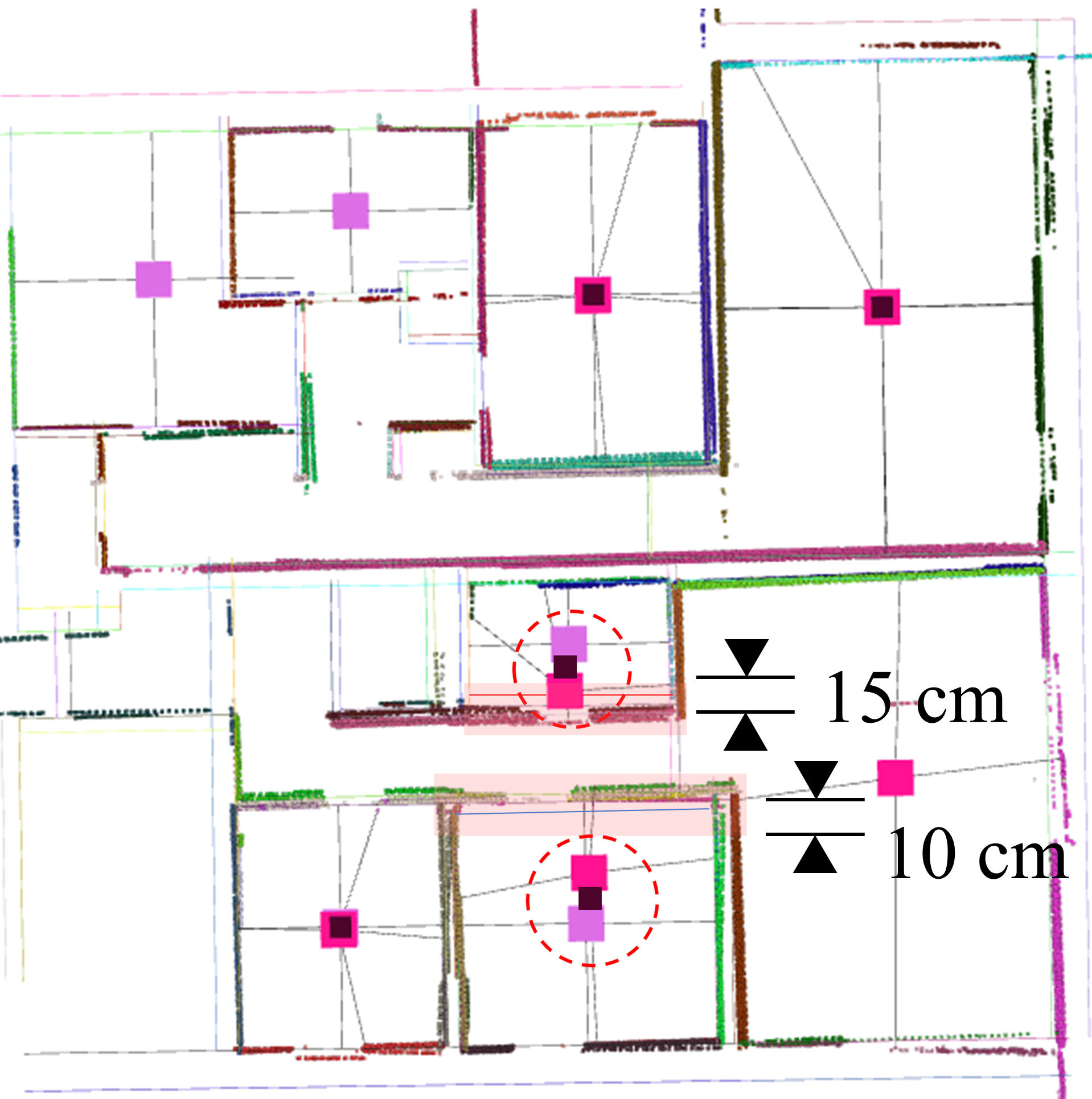}
\caption{RE4}
\label{fig:RE4}
\end{subfigure}
\caption{ \textit{diS-Graphs} of real construction sites showing detected deviations. Red dotted circles indicate room deviations, red-shaded translucent rectangles highlight wall deviations, and black markers represent deviation factors. See in (a) a zoom-in of a room deviation (non-coincident room centers, represented by pink and purple squares) and of a wall-surface deviation.
}
\label{fig:isgraphs_plus_real_deviations}
\end{figure*}
\section{Graph Optimization}
\label{sec:graph_optimization}
Before the \textit{S-Graph} and \textit{A-Graph} are coupled, the \textit{S-Graph} state at time $t$ can be defined as~\cite{bavle_sgraphs_2023}: 
~
\begin{equation}
\label{eq:s_graph_only_state}
 \mathbf{s_{1_\mathcal{S}}} = [ {\mathbf{x}}^\mathcal{S}_{{r}_i},\ {\boldsymbol{{\boldsymbol{\pi}}}}^\mathcal{S}_{s_l}, \ {\boldsymbol{\rho}}^\mathcal{S}_{s_p}, \ {\boldsymbol{\kappa}}^\mathcal{S}_{s_q}, \ {\boldsymbol{\xi}}^\mathcal{S}_{s}, \ {\mathbf{x}}^\mathcal{S}_{O}]^\top
\end{equation}
Similarly, the \textit{A-Graph} state is defined as~\cite{shaheer_graph-based_2023}:
\begin{equation}
\label{eq:a_graph_only_state}
\mathbf{s_{1_\mathscr{A}}} = [{\boldsymbol{\pi}}^\mathscr{A}_{a_j}, \ {\boldsymbol{\rho}}^\mathscr{A}_{a_n}, \ {\boldsymbol{\mathcal{D}}}^\mathscr{A}_{a_o},\ {\boldsymbol{\xi}}^\mathscr{A}_{a}]^\top
\end{equation}
\noindent where 
$^\mathcal{S}\mathbf{x}_{O}$ models the drift between the odometry frame $O$ and the \textit{S-Graph} reference frame $\mathcal{S}$, and the other variables are described in Section \ref{sec:proposed_approach}. Note, if at time $t$ there is no match between the \textit{A-Graph} and \textit{S-Graph}, we perform single \textit{S-Graph} optimization as detailed in \cite{bavle_sgraphs_2023} until we detect it. 

\textbf{Alternating Optimization.} \label{sec:alternating_optimization}
Alternating optimization is further performed in two steps as follows:
 
\textbf{\textit{Transformation Estimation:}} Upon receiving the match (Section~\ref{sec:graph_matching}) and performing graph 
coupling (Section~\ref{sec:data_association}), we combine Eq. \ref{eq:s_graph_only_state} and Eq. \ref{eq:a_graph_only_state} to make a global state with additional transformation $\mathbf{s_2} = [\mathbf{s_{1_\mathcal{S}}}, \, \mathbf{s_{1_\mathscr{A}}} \,  \mathbf{p}^\mathscr{A}_{\mathcal{S}}]$.
It is important to note that at this stage, the deviated wall-surface and room are not included for estimating $\mathbf{p}^\mathscr{A}_{\mathcal{S}}$.

\textbf{\textit{Deviation Estimation:}} After optimizing $\mathbf{s_2}$ we already have an initial guess of the transformation between the \textit{A-Graph} and the \textit{S-Graph}, and we can incorporate the deviated wall-surface and room entities into the graph with appropriate deviation factors. Our state then becomes $\mathbf{s_3} = [\mathbf{s_2}, \ \mathbf{d}^{^{}{\boldsymbol{\pi}}_{a_i}}_{^{}{\boldsymbol{\pi}}_{s_i}}, \ \mathbf{d}^{{}^{}{\boldsymbol{\rho}}_{a_i}}_{^{}{\boldsymbol{\rho}}_{s_i}}]$ 
When optimizing $\mathbf{s_3}$ we keep $\mathbf{s_2}$ constant to obtain a good initial estimation of the deviation between the matched deviated entities. 

\textbf{Joint Optimization.}
\label{sec:combined_optimization}
Finally, after getting the initial estimates of the transformation between the origins and the deviations between the semantic entities, we optimize the whole state $\mathbf{s_3}$ to simultaneously estimate the position of each semantic entity, deviations, and the transformation between the two graphs. 

\section{Experimental Evaluation}
\label{sec:experimental_evaluation}
\subsection{Methodology}
\textbf{Setup.} We evaluated our algorithm in both simulated and real environments. Both types of experiments were performed using a laptop computer with an Intel i9-11950H (8 cores, 2.6 GHz) with 32 GB of RAM. 

\textbf{Baselines}.
We have selected the following LiDAR-based baselines for comparisons due to their suitability, their reported results, and the availability of their code: \textit{AMCL} \cite{dieter_fox_monte_2001}, \textit{UKFL} \cite{koide_portable_2019}, \textit{OGM2PGM} \cite{torres_occupancy_2023},  \textit{IR-MCL} \cite{kuang_ir-mcl_2023}, and \textit{iS-Graphs} \cite{shaheer_graph-based_2023}. 
Although other works do localization in imprecise architectural plans, the absence of open-source implementations prevents direct comparison with these methods. As each selected baseline takes a different map input for localization, we used 2D occupancy grid maps for \textit{AMCL} and \textit{OGM2PGM}, 3D meshes for \textit{UKFL} and \textit{IR-MCL}, and \textit{A-Graphs} for \textit{iS-Graphs} and our method \textit{diS-Graphs}, respectively. 

\textbf{Simulated Datasets}. 
We validate the algorithms on five simulated datasets (\textit{SE1} to \textit{SE5}) using Gazebo to recreate the robot, LiDAR sensor, and 3D environments from actual architectural plans. We report absolute trajectory error (ATE) and localization convergence success rates.

\textbf{Real Datasets.}
We collected data using a legged robot with Velodyne VLP-16 LiDAR at five construction sites (\textit{RE1} to \textit{RE5}) with existing architectural plans. We report Root Mean Square Error (RMSE) between estimated 3D maps and ground truth plans (indicating localization accuracy since mapping quality reflects pose estimation), plus convergence rates and computation times.

\textbf{Deviations.}
Note that, given the absence of actual ground truth deviations in a real environment with respect to the plans, we explicitly deviate the wall-surfaces in the architectural plans to have a ground truth estimate of the deviations for both simulated and real datasets. For all datasets, we performed five separate tests introducing uniform random wall-surface deviations in architectural plans ranging from $5 \ \textrm{cm}$ to $40 \ \textrm{cm}$ in translation, and $5^{\circ}$ and $15^{\circ}$ in rotations. 
Moreover, to test the robustness of our algorithm against deviations, we conducted 25 experiments per dataset with varying deviations. 

\textbf{{Ablations.}}
We conducted ablation studies on two key components of our algorithm. 1.~\textbf{\textit{Covariance assignment:}} As mentioned in the graph matching and graph merging (Section.~\ref{sec:graph_matching} \& \ref{sec:data_association}), we assign covariances to the deviation factors of the semantic elements, based on their deviation likelihood. Here, we analyze the effect of assigning equal covariances, referred to as uniform covariances (UC), to both deviated and non-deviated elements.
 2.~\textbf{\textit{Alternating optimization:}} In graph optimization (Section.~\ref{sec:alternating_optimization}) we discuss the use of alternating optimization for simultaneous localization and deviation estimation. Here, we study the performance when using only a single optimization cycle (SO).
\subsection{Results and Discussion} 
\label{sec:results_and_discussion}
\begin{figure}[!b]
\begin{subfigure}{0.24\textwidth}
 \centering
\includegraphics[width=0.9\textwidth]{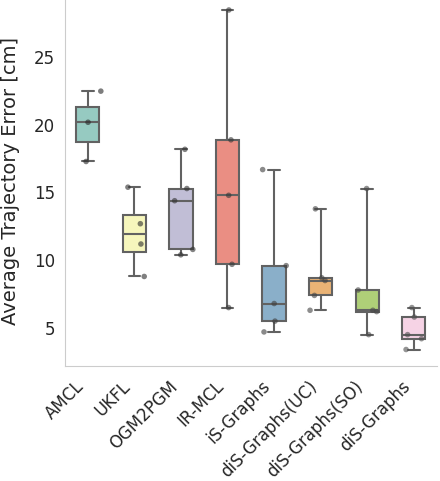}
\caption{ATE Comparison}
\label{fig:ate_plot}
\end{subfigure}
\begin{subfigure}{0.24\textwidth}
\includegraphics[width=0.9\textwidth]{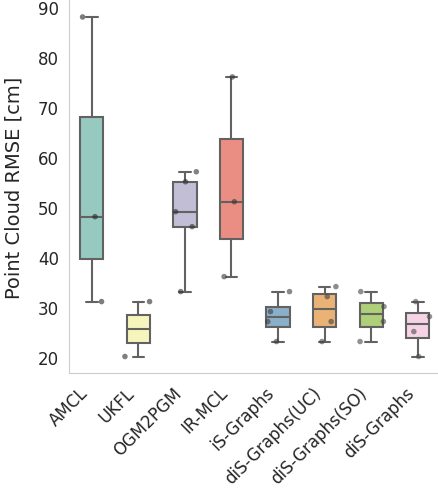}
\caption{RMSE Comparison}
\label{fig:rmse_plot}
\end{subfigure}
\caption{\textbf{a)} Distribution of Average Trajectory Error (ATE) in simulated datasets. \textbf{b)} Distribution of point cloud alignment error (RMSE) in real datasets. 
UKFL's only 2 convergent cases prevent fair comparison.
}
\label{fig:plots}
\end{figure}

\textbf{Absolute Trajectory Error.}
Table \ref{tab:ate_simulated_exp} shows the average ATE for all baselines and our \textit{diS-Graphs} in the presence of deviations between ``as-planned'' and ``as-built'' environments. Our method
shows an error reduction of around $75.8\%$ compared to AMCL, $64.9\%$ compared to OGM2PGM, $69 \%$ compared to IR-MCL, $47.2 \%$ compared to UKFL, and a reduction of $43\%$ over \textit{iS-Graphs}.
\rowcolors{5}{gray!25}{white}
\begin{table}[!b]
\setlength{\tabcolsep}{5pt}
\caption{
Average ATE [cm] for simulated experiments. Each entry represents the mean of 5 tests with different amounts of deviation.
\textbf{Bold} values are the best and the second best are underlined. `-' refers to an unsuccessful run.}
 \scriptsize
\centering
 \ra{1.0}
\begin{tabular}{l|c c c c c|c}
\toprule
 & \multicolumn{5}{c}{\textbf{Dataset}} \\ 
\toprule
\textbf{Method} & \multicolumn{6}{c}{\textbf{ATE [cm]}} \\ \toprule
 & \textit{SE1} & \textit{SE2} & \textit{SE3}  &\textit{SE4} & \textit{SE5} & Avg. \\ 
   \midrule
  AMCL \cite{dieter_fox_monte_2001} &    $-$  &  17.2 &     $-$  &  20.1 & 22.4 & 19.9 \\ 

  UKFL \cite{koide_portable_2019} &    12.6 &  15.3 &    $-$     &  8.7&   11.1  & 9.1\\ 

OGM2PGM \cite{torres_occupancy_2023} &  15.2 &  18.1  &   10.7 &  10.3 & 14.3 & 13.7\\ 

IR-MCL \cite{kuang_ir-mcl_2023}&    14.7 &  \underline{6.4}  &   \underline{9.6} &  28.4 &   18.8 & 15.5\\

\textit{iS-Graphs} \cite{shaheer_graph-based_2023} &  \underline{5.4} &   6.7&   16.6  &  \underline{4.6} &      \underline{9.5} & \underline{8.5}\\

\midrule
\textit{diS-Graphs} (UC) &  6.2&  7.3 &  13.7   & 8.4&    8.6  & 8.8\\

\textit{diS-Graphs} (SO) &4.4 &  6.1 &  15.2  & 7.7  &   6.2 & 7.9\\
\midrule

\textit{diS-Graphs} (Ours) & \textbf{3.3} & \textbf{4.1} & \textbf{6.4}  & \textbf{4.4} & \textbf{ 5.7 } & \textbf{4.8}\\
\bottomrule
\end{tabular}
\label{tab:ate_simulated_exp}
\end{table}
Fig. \ref{fig:ate_plot} summarizes the ATE performance of all the baseline algorithms. AMCL shows the highest median ATE and a relatively narrow distribution, indicating consistently high error rates. UKFL and OGM2PGM demonstrate moderate performance with similar median ATEs, although OGM2PGM shows a wider range of errors. IR-MCL exhibits the largest variability, suggesting inconsistent performance in different scenarios. 
AMCL, OGM2PGM, IR-MCL, UKFL and \textit{iS-Graphs} assume no deviations between ``as-planned'' and ``as-built'' environments. These methods attempt to match laser scans with an incorrect reference map, leading to consistent misalignment. Therefore, they show higher ATEs across datasets, indicating their vulnerability to architectural deviations. Our \textit{diS-Graphs} can detect and estimate the deviations between  ``as-planned'' and ``as-built'' environments, resulting in the lowest ATE.

\textbf{Point Cloud Alignment Error.}
Table \ref{tab:rmse_real_data} shows the RMSE of the point clouds with respect to the ground truth for all baseline methods and ours. In the presence of deviations between ``as-planned'' and ``as-built'', \textit{diS-Graphs} shows $53.5\%$ better accuracy than AMCL, $45.8\%$ better than OGM2PGM, $51.8\%$ better than IR-MCL, and $7\%$ better than \textit{iS-Graphs}. Although UKFL's average error is equal to \textit{diS-Graphs}', it has a very low convergence rate (see Table \ref{tab:convergence_rate}) rendering the comparison unfair. Moreover, it cannot estimate the deviations between ``as-planned'' and ``as-built'' environments. Fig. \ref{fig:rmse_plot} summarizes the performance of all algorithms in real environments. AMCL exhibits the highest median and widest interquartile range, indicating greater variability in performance.  IR-MCL presents a large spread of results, while OGM2PGM shows moderate performance with a smaller range of variability compared to AMCL and IR-MCL. 
Although UKFL and \textit{diS-Graphs} show the lowest median RMSE, suggesting superior accuracy, the comparison with UKFL is not fair because of its low convergence rate.
 Because of our simultaneous estimation of deviations and initial transformation, we can not only simultaneously globally localize the robot but also estimate the deviations between semantic elements of ``as-planned'' and ``as-built'' environments, moreover improving the overall accuracy compared to other algorithms. 
\rowcolors{4}{gray!23}{white}
\begin{table}[!ht]
\setlength{\tabcolsep}{5pt}
\caption{Point cloud RMSE [cm] for real experiments. 
\textbf{Bold} values are the best and the second best are underlined. `-' refers to an unsuccessful run.}
\scriptsize
\centering
\ra{1.0}
\begin{threeparttable}
\begin{tabular}{l|c c c c c|c}
\toprule
& \multicolumn{5}{c}{\textbf{Dataset} } \\
\toprule
\textbf{Method} & \multicolumn{5}{c}{\textbf{Point Cloud RMSE} [cm]} \\
\toprule
   & \textit{RE1} &  \textit{RE2}  & \textit{RE3} & \textit{RE4} & \textit{RE5} & Avg \\ 
\midrule
AMCL\cite{dieter_fox_monte_2001}  & 0.48 & 0.88  &  0.31 & $-$ &$-$ & 0.56\\ 

UKFL \cite{koide_portable_2019} & 0.31 & $-$ & \textbf{0.20}   & $-$ &  $-$ & 0.26*\\ 
OGM2PGM\cite{torres_occupancy_2023}  & 0.46 & 0.57  & 0.33  & 0.49 &\textbf{0.55} &  0.48 \\ 
IR-MCL\cite{kuang_ir-mcl_2023}   & 0.51 & 0.76 & 0.36 & $-$  & $-$ & 0.54 \\
\textit{iS-Graphs} \cite{shaheer_graph-based_2023}  & \underline{0.27} & \underline{0.29} &  \underline{0.23} & \underline{0.33}  & $-$  & \underline{0.28}\\
\midrule
\mbox{\textit{diS-Graphs} (UC)}  & 0.27 & 0.32 & 0.23  & 0.34 & $-$ & 0.29\\ 
\mbox{\textit{diS-Graphs} (SO)} & 0.27 & 0.30 & 0.23 & 0.33 & $-$ & {0.28}\\
\midrule
\textit{diS-Graphs (ours)} & \textbf{0.25} & \textbf{0.28}  & \textbf{0.20}  & \textbf{0.31} & $-$  & \textbf{0.26}\\ 
\bottomrule 
\end{tabular}

\begin{tablenotes}
      \small
      \item * Omitted due to low convergence. 
    \end{tablenotes}
\end{threeparttable}
\label{tab:rmse_real_data}
\end{table}
\begin{figure*}[t!]
\centering
\begin{subfigure}{0.17\textwidth}
 \centering
\includegraphics[width=1\textwidth]{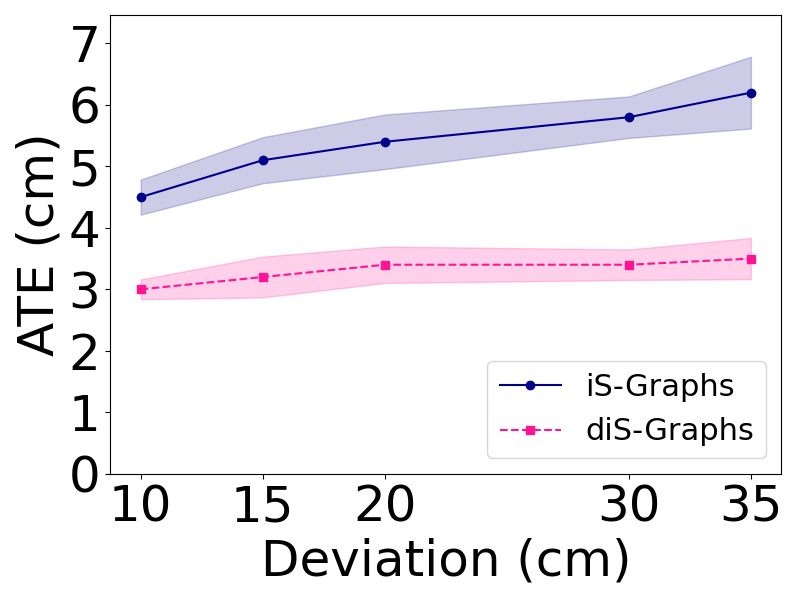}
\caption{SE1}
\label{fig:SE1}
\end{subfigure}
\begin{subfigure}{0.17\textwidth}
\centering
\includegraphics[width=1\textwidth]{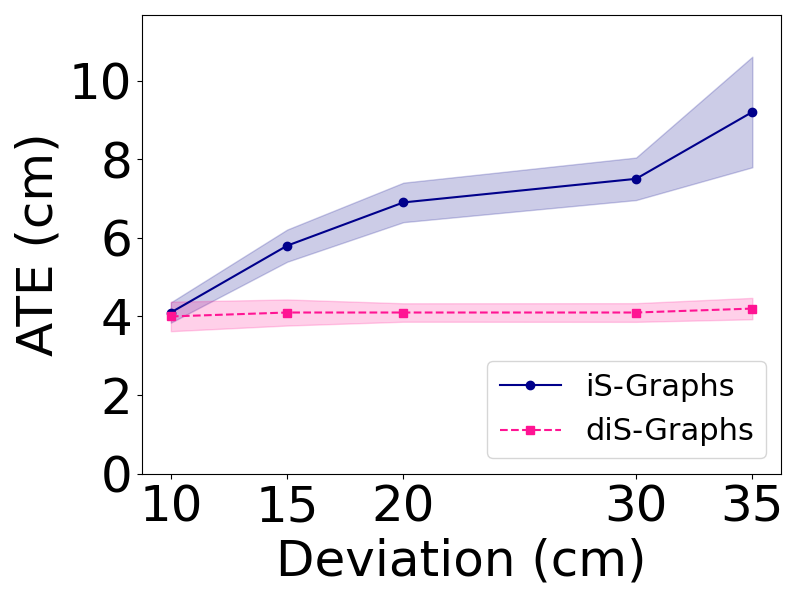}
\caption{SE2}
\label{fig:SE2}
\end{subfigure}
\begin{subfigure}{0.17\textwidth}
\centering
\includegraphics[width=1\textwidth]{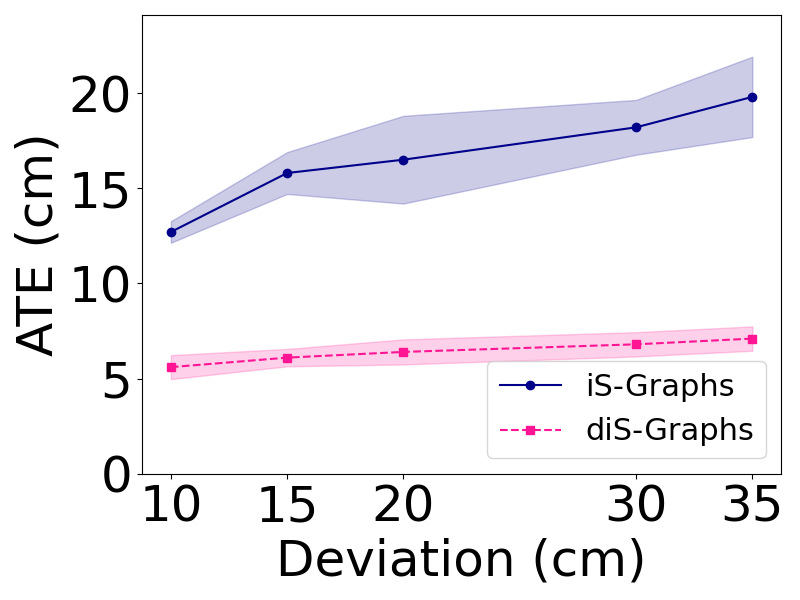}
\caption{SE3}
\label{fig:SE3}
\end{subfigure}
\begin{subfigure}{0.17\textwidth}
\centering
\includegraphics[width=1\textwidth]{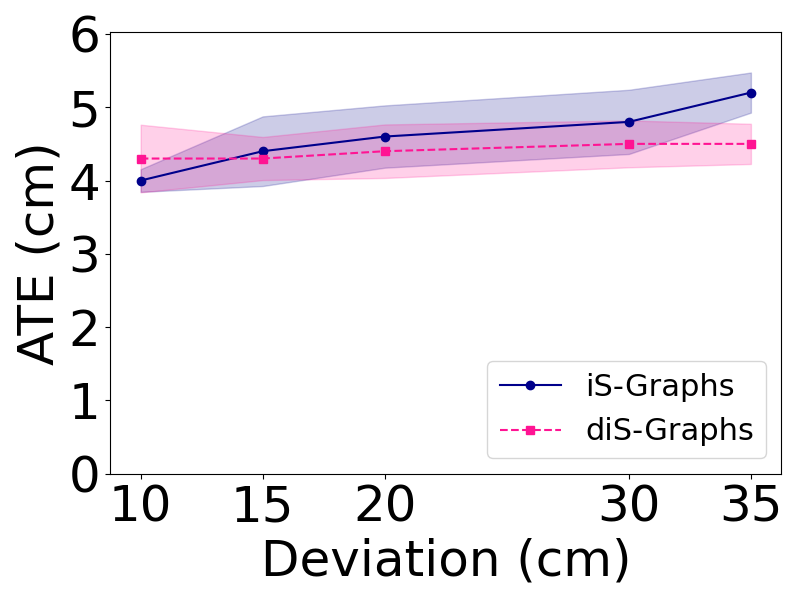}
\caption{SE4}
\label{fig:SE4}
\end{subfigure}
 \centering
\begin{subfigure}{0.17\textwidth}

\includegraphics[width=1\textwidth]{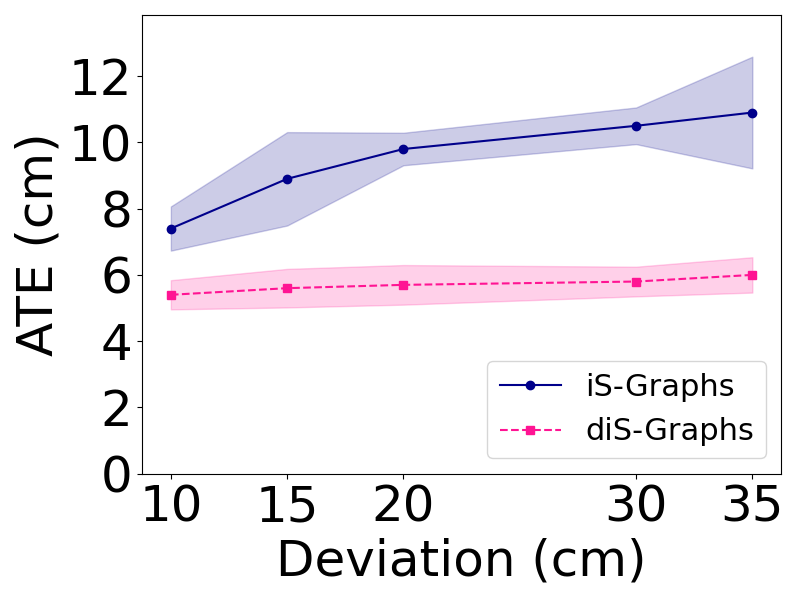}
\caption{SE5}
\label{fig:SE5}
\end{subfigure}
\caption{
Comparison of Average Trajectory Error (ATE) against the amount of deviation in simulated datasets.
}
\label{fig:ate_vs_dev}
\end{figure*}

\textbf{Convergence Rate.}
Our approach demonstrates superior convergence rates across both simulated and real environments (Table \ref{tab:convergence_rate}). In simulations, our method achieves a convergence rate $60\%$ better than AMCL, $44\%$ better than UKFL, and $12\%$ better than OGM2PGM. While our convergence rate matches that of IR-MCL in simulations, IR-MCL exhibits inconsistent performance in terms of ATE and requires retraining for each new dataset. Similarly, in real environments (\textit{RE1-RE4}), our method maintains the highest convergence rate among all baselines, leveraging its ability to detect deviated elements in the environment. The only exception is in \textit{RE5}, where our algorithm fails due to insufficient room detection for effective graph matching, because of noisy sensor data. Moreover, AMCL and UKFL's heavy reliance on accurate initial position estimates leads to degraded convergence rates in environments with complex geometries. OGM2PGM's accuracy suffers particularly in symmetric environments. Our method overcomes these limitations through its robust deviation detection capabilities, though it requires a sufficient number of distinguishable rooms to function effectively.
\rowcolors{4}{gray!25}{white}
\begin{table}[!h]
\setlength{\tabcolsep}{4pt}
\caption{Convergence rate [\%] of simulated and real experiments.}
\centering

\scriptsize
\ra{1.0}
\begin{tabular}{l | c c c c c | c c c c c}
\toprule
& \multicolumn{10}{c}{\textbf{Dataset}} \\ 
\midrule
& \multicolumn{10}{c}{\textbf{Convergence Rate [\%]}} \\ 
\midrule
\textbf{Method} & \textit{SE1} & \textit{SE2}  & \textit{SE3} & \textit{SE4} & \textit{SE5} & \textit{RE1} & \textit{RE2}  & \textit{RE3} & \textit{RE4} & \textit{RE5} \\ \midrule
AMCL \cite{dieter_fox_monte_2001}   & 0 & 100  & 0 & 80& 20&  80 & 100 & 100 & 0  & 0  \\
UKFL \cite{koide_portable_2019}   & 80& 80   & 0  & 40   & 80   & 60 & 0 & 60 & 0 & 0 \\ 
OGM2PGM \cite{torres_occupancy_2023} & 80 & 100 & 100 & 80 & 80 & 60 & 80 & 100 & 60 & 80  \\
IR-MCL \cite{kuang_ir-mcl_2023} &100 & 100 & 100 & 20 & 100 & 60 & 100 & 20 & 0 & 0 \\
\textit{iS-Graphs} \cite{shaheer_graph-based_2023} &  60    &   20    &   100    &   80   &   100     &  60     &   20    &   40    & 60 &  0   \\
\textit{diS-Graphs}  &  100    &   100  &  100    &   100    &  100  & 100 & 100 &100 & 100  & 0 \\
\bottomrule
\end{tabular}
\label{tab:convergence_rate}
\end{table}
  
\rowcolors{5}{gray!23}{white}
\begin{table}[!h]
\setlength{\tabcolsep}{3pt}
\caption{Convergence time [s] and computation time [ms] on real experiments. \textbf{Bold} values are the best and the second best are underlined. `-' refers to an unsuccessful run.}
\centering
\ra{1.0}
\scriptsize
\begin{tabular}{l | c c c c c | c c c c c}
\toprule
& \multicolumn{10}{c}{\textbf{Dataset}} \\ 
\midrule
 & \multicolumn{5}{c}{\textbf{Convergence Time} [s]} & \multicolumn{5}{c}{\textbf{Computation Time} [ms]} \\
\midrule
\textbf{Method} & \textit{RE1} & \textit{RE2}  & \textit{RE3} & \textit{RE4} & \textit{RE5} & \textit{RE1} & \textit{RE2}  & \textit{RE3} & \textit{RE4} & \textit{RE5} \\ \midrule
AMCL \cite{dieter_fox_monte_2001}   & \underline{24} & 89  & 29 & $-$& $-$& \textbf{2}  & \textbf{2} & \textbf{2}  & $-$  & $-$  \\
UKFL \cite{koide_portable_2019}   & 126& $-$   & \textbf{8}  & $-$   & $-$   & 104& $-$ & 119& $-$ & $-$ \\ 
OGM2PGM \cite{torres_occupancy_2023} & \textbf{16} & \underline{38} & 27 & \textbf{35} & \textbf{79} & \textbf{2}  & \textbf{2}  & \textbf{2}  & \textbf{2} & \textbf{2}  \\
IR-MCL \cite{kuang_ir-mcl_2023} &\textbf{16} & \textbf{26} & \underline{13} & $-$ & $-$ & 92 & 90 & 89 & $-$ & $-$ \\
\textit{iS-Graphs} \cite{shaheer_graph-based_2023} &  155    &   101    &   78    &   \underline{139}    &   $-$     &   57    & 78      &   78    &    \underline{64}   &  $-$   \\
\textit{diS-Graphs}  &  81    &   43    &   46    &   \underline{139}    &   $-$  & \underline{56} & \underline{77} &\underline{76} &70  & $-$ \\
\midrule
\textbf{Seq. Len.} [s] & 657 & 170 & 488 & 657 & 559  &  657 & 170 & 488 & 657 & 559\\
\bottomrule
\end{tabular}
\label{tab:compute_time}
\end{table}
\rowcolors{5}{gray!23}{white} 
\textbf{Convergence and Computation Time.} Table \ref{tab:compute_time} shows the convergence time for each algorithm. The convergence time is the time it takes the algorithm to globally localize. IR-MCL and OGM2PGM have the best convergence time. \textit{iS-Graphs} and \textit{diS-Graphs} have considerably longer convergence times because they need to detect a certain number of rooms in the environment for graph matching to find a unique match. The graph matching of \cite{shaheer_graph-based_2023} struggles to resolve symmetries during the matching process, resulting in longer convergence times. However, the modifications in the graph matching algorithm (Section \ref{sec:graph_matching}) proposed in this work improve the symmetry resolution ability and reduce the convergence time by almost $20 \%$. Table \ref{tab:compute_time} shows the computation time for each algorithm. We specifically report the time spent for each pose update. On average, the computation time of our algorithm is $70$ milliseconds per pose update, showing real-time performance. Our computation time is the best compared to other 3D algorithms.
OGM2PGM and AMCL. The smaller computation time of the others is due to their simpler optimization in 2D. 

\begin{figure}[!h]
\begin{subfigure}{0.24\textwidth}
 \centering
\includegraphics[width=0.8\textwidth]{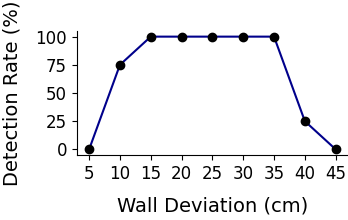}
\caption{Translational Deviation}
\label{fig:dev_plot}
\end{subfigure}
\begin{subfigure}{0.24\textwidth}
\includegraphics[width=0.8\textwidth]{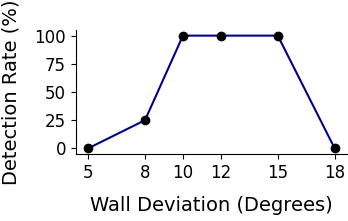}
\caption{Rotational Deviation}
\label{fig:angle_plot}
\end{subfigure}
\caption{Average deviation detection rate for real datasets.}
\label{fig:deviation_plots}
\end{figure}

\textbf{Deviation Estimation.} Fig. \ref{fig:deviation_plots} shows the amount of deviation our \textit{diS-Graphs} can correctly estimate in real environments. 
The minimum detectable deviation is bounded by the LiDAR resolution. With our LiDAR errors having a stamdard deviation of $3\ cm$ \cite{glennie_calibration_2016}, and considering that $99.7\%$ of measurements fall within three standard deviations, our \textit{diS-Graph} can reliably detect deviations only when they exceed $9\ cm$.
The maximum translational and rotational wall-surface deviation that our \textit{diS-Graphs} can detect is $35\ cm$ and $15^\circ$ respectively. Fig. \ref{fig:isgraphs_plus_real_deviations} shows several qualitative results from our real experiments. The robot can successfully localize, map, and estimate deviations in these environments. Note that we only show the rooms and walls for better understanding, and all the other semantic elements and the robot are not shown. We do not show \textit{RE5} in Fig. \ref{fig:isgraphs_plus_real_deviations}, as the robot could not localize itself in this sequence.  

\textbf{Robustness against Deviations.} Figure \ref{fig:ate_vs_dev} shows the effect of deviations on the localization performance of \textit{iS-Graphs} and \textit{diS-Graphs}. The results consistently show that \textit{diS-Graphs}, which incorporates deviation detection and modeling, maintains a relatively stable ATE even as the deviation increases from  $10$ cm to $35$ cm. In contrast, \textit{iS-Graphs}, which lacks explicit deviation modeling capabilities, exhibits a clear upward trend in ATE as deviations increase. 

\textbf{Ablation Study.} Table \ref{tab:ate_simulated_exp} shows that associating `uniform covariance' in simulated datasets (\textit{diS-Graphs} (UC)) the algorithm cannot differentiate between deviated and non-deviated elements which results in poor pose estimation. 
Fig. \ref{fig:ate_vs_dev} demonstrates that \textit{iS-Graphs} with uniform covariance shows higher error variability compared to our method which uses deviation probability-based covariance assignment.
In some cases, the use of uniform covariances results in even worse performance than \textit{iS-Graphs}. Similarly, when using single optimization (\textit{diS-Graphs} (SO)) instead of alternating optimization, the algorithm cannot differentiate between the transformation between two graphs and the deviations between their elements, resulting in higher ATE as shown in Table \ref{tab:ate_simulated_exp}. Table \ref{tab:rmse_real_data} shows the ablation of uniform covariances and single optimization in real-world datasets. Using uniform covariances (\textit{diS-Graphs} (UC)) and single optimization (\textit{diS-Graphs} (SO)) results in lower mapping accuracy due to the algorithm's inability to accurately map the deviated elements and differentiate between initial transformation and deviations simultaneously.
\section{Conclusions and Future Work}
\label{sec:conclusion}
This paper presents a novel approach to tightly couple SLAM with imprecise architectural plans. Our algorithm establishes direct correspondences between structural elements in ``as-built" environments and their ``as-planned" counterparts, enabling simultaneous robot localization and deviation estimation between the two representations. 
By detecting and estimating the deviations between ``as-built'' and ``as-planned'' environments through tight coupling, our method outperforms the current best approach with 43\% better localization in simulations, 7\% improved mapping accuracy in real environments, and enhanced robustness to architectural deviations. Our algorithm provides an estimate of existing deviations up to $35 \ cm$ in translation and $15^{\circ}$ in rotation. 
Our algorithm is limited by the need for sufficient distinctive semantic elements (i.e., wall-surfaces and rooms) to provide unique matches. As future work, we plan to detect and match more semantic elements and support rooms with non-rectangular geometries, including curved walls and partial floorplans. Additionally, we aim to integrate multi-modal sensing (RGB-D, vision-based semantic segmentation) to enhance robustness in visually degraded environments. These improvements will enhance the convergence rate and enable deviation detection for additional detected entities beyond just rooms and walls.

\balance
\bibliographystyle{IEEEtran}
\bibliography{Bibliography}

\end{document}